\documentclass{article} 
\usepackage{sections/fsmem_preprint,times}

\usepackage{amsmath,amsfonts,bm}

\def\eqref#1{equation~\ref{#1}}

\def\1{\bm{1}}

\DeclareMathAlphabet{\mathsfit}{\encodingdefault}{\sfdefault}{m}{sl}
\SetMathAlphabet{\mathsfit}{bold}{\encodingdefault}{\sfdefault}{bx}{n}

\usepackage{hyperref}
\hypersetup{hidelinks}
\usepackage{url}
\usepackage{amsmath,amssymb}
\usepackage{amsthm}
\theoremstyle{definition}
\newtheorem{definition}{Definition}
\usepackage{booktabs}
\usepackage{array,multirow}
\usepackage{algorithm}
\usepackage[noend]{algpseudocode}
\algrenewcommand{\algorithmicrequire}{\textbf{Input:}}
\algrenewcommand{\algorithmicensure}{\textbf{Output:}}
\usepackage{graphicx}
\usepackage{wrapfig}
\usepackage{needspace}
\usepackage{xcolor}
\DeclareRobustCommand{\rev}[1]{#1}
\usepackage{colortbl}
\usepackage{tcolorbox}
\newtcolorbox[
  auto counter,
  number within=section
]{takeaway}{
  colback=orange!2,
  colframe=orange!35!gray,
  boxrule=0.4pt,
  arc=1pt,
  left=3pt,
  right=3pt,
  top=3pt,
  bottom=3pt,
  before upper={\textbf{Takeaway~\thetcbcounter}\ }
}
\makeatletter
\newenvironment{figurehere}{%
  \par\medskip\noindent\begin{minipage}{\textwidth}%
  \def\@captype{figure}\centering
}{\end{minipage}\par\medskip}

\makeatother

\usepackage{enumitem}
\newcommand{\Ent}{\mathcal{E}}
\newcommand{\Act}{\mathcal{A}}
\newcommand{\St}{\mathcal{S}}

\title{Internalized Experience: How Training Data Shapes Memory Formation in LLMs}

\author{Xianxuan Long, Xinnan Dai, Kai Guo, and Jiliang Tang \\
Michigan State University}

\begin{document}
\raggedbottom

\maketitle

\begin{abstract}
Parametric memory enables large language models (LLMs) to retain information from past experience and reuse it in subsequent tasks. Yet how training data supports the internalization of reusable task rules remains unclear. We introduce \textbf{FSMem} (Finite-State Machine Memory), a state-transition framework for constructing training data and studying how data content and organization shape memory formation. We evaluate rule reuse across synthetic worlds, classical planning, and code execution through valid-action identification and execution-outcome prediction, complemented by next-operation prediction in code.
Our experiments reveal several systematic learning patterns. Under matched training conditions, even unchanged task rules become harder to internalize when learned alongside a larger variety of action types; by contrast, scaling the training data to provide broader state--action coverage generally improves rule learning. We further find that training at a single execution depth produces depth-specialized behavior, with models performing best when training and test executions have similar lengths. Motivated by this finding, we construct cumulative corpora that combine short and long trajectories, which can outperform long-trajectory-only training while using fewer training tokens. Together, these results demonstrate that the same task rules can be internalized differently depending on how execution experience is constructed, and that understanding these learning patterns can guide the design of more effective training data for parametric memory.
\end{abstract}

\section{Introduction}
\label{sec:intro}
Large language models (LLMs) are expected to build on past experience when facing new tasks and contexts. Memory supports this reuse by preserving information beyond the context in which it was acquired~\citep{wang2023longmem}. Existing approaches realize this capacity mainly through text-based memory, which stores information externally for later retrieval~\citep{lewis2020rag,zhong2024memorybank,lu2023memochat}, or parametric memory, which incorporates information into model parameters through updates~\citep{roberts2020much,meng2022rome,meng2023memit,back2026loramemory}.
In this work, we focus on parametric memory given its potential to turn past experience into lasting improvements in model capabilities.

\begin{figure}[t]
\centering
\includegraphics[width=0.85\textwidth]{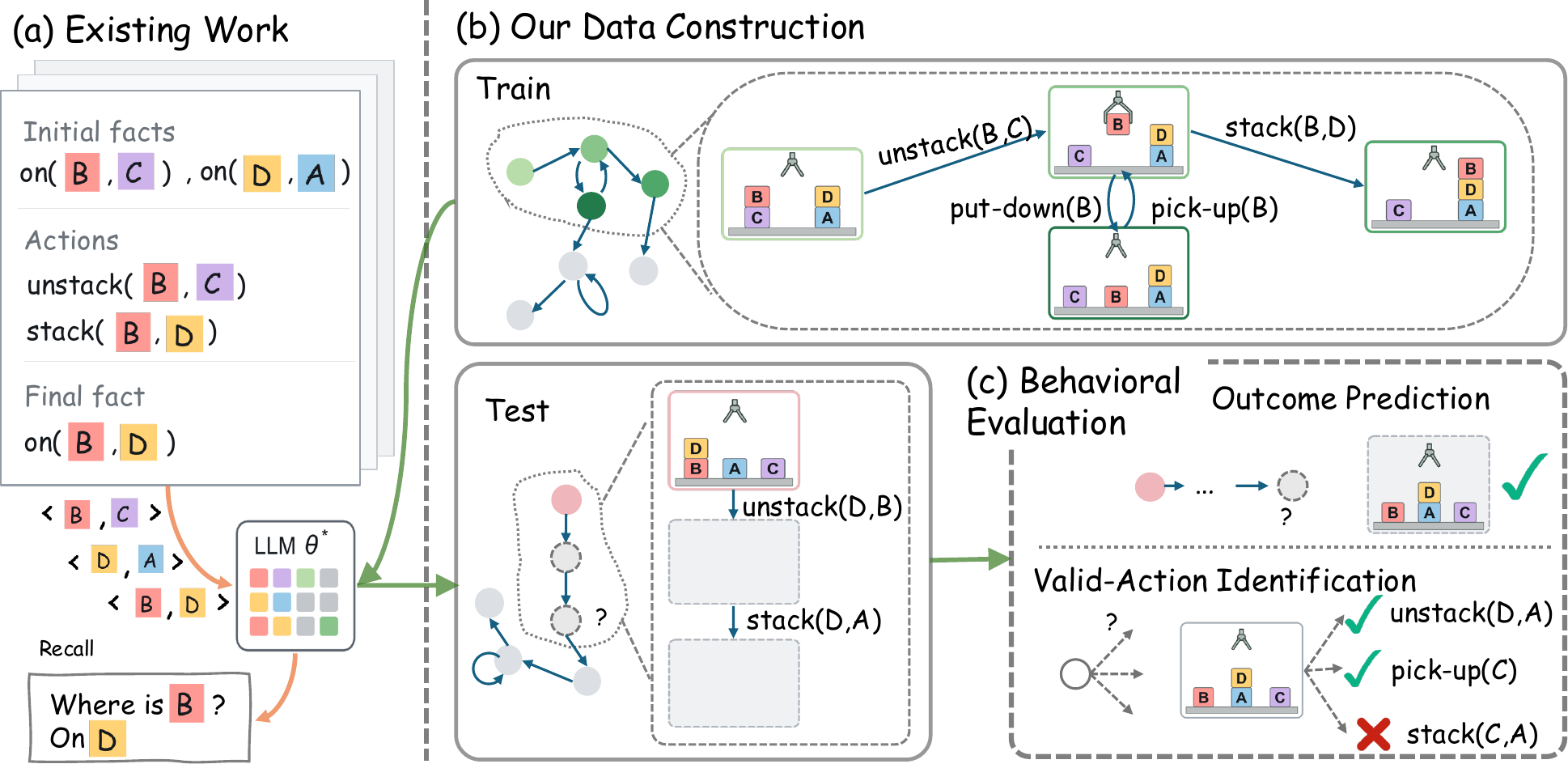}
\vspace{-6pt}
\caption{An illustration of FSMem in Blocksworld.
(a)~Recalling facts about specific entities.
(b)~Constructing training data from execution trajectories.
(c)~Identifying valid actions and predicting execution outcomes on held-out
instances, without access to training records.}
\label{fig:framework}
\end{figure}

Existing research on LLMs' parametric memory focuses on mechanisms for encoding useful information into model parameters~\citep{de2021editing,mitchell2021fast,charakorn2026doctolora} or dedicated parametric modules for later use~\citep{lample2019large,wang2024wiserethinkingknowledgememory,behrouz2025titans,wei2026mlpmemoryretrieverpretrainedmemory}. These studies have advanced our understanding of memory capacity~\citep{allenzhu2024physics33,xu2026loraremembersparametricmemory}, the internalization of factual information~\citep{petroni2019language,allenzhu2024physics31,chang2024acquire}, and reasoning over learned information~\citep{allen2025physics,wang2024grokked,back2026loramemory}. What models internalize also depends on the training data itself~\citep{tirumala2022memorization,ye2026cramless}. For example, planning trajectories~\citep{kokel2025acpbench} and program execution traces~\citep{nye2021scratchpad} have been used to study how LLMs reason about task execution. These records contain information about particular entities and their states, as illustrated in Figure~\ref{fig:framework}(a). However, existing work has overlooked the fact that execution trajectories involving different entities could follow the same task rules: when an action is valid and how it changes the state, and apply across different entities and states. They are reusable across instances of a given task~\citep{fikes1971strips,ghallab1998pddl,yang2007learning}. How the task content and organization of these records facilitate the internalization of such reusable rules remains less well understood.

Studying the internalization of task rules requires distinguishing which rules training data contains from how their executions are recorded. In this work, we introduce \textbf{FSMem} (\textbf{F}inite-\textbf{S}tate \textbf{M}achine \textbf{Mem}ory), a framework that makes this distinction explicitly by representing task execution as a finite-state machine (Section~\ref{sec:formulation}). For the Blocksworld example in Figure~\ref{fig:framework}(b), states describe the configuration of blocks and the gripper's status, while concrete actions manipulate particular blocks. The transition function specifies whether an action is valid and, if so, the resulting state; trajectories record sequences of concrete actions and states. This representation allows us to vary the action types selected for supervision and, with the rules fixed, the state–action coverage and depth of execution records.

Consistent with our understanding of parametric memory, past experience should induce lasting changes in model parameters that influence subsequent behavior~\citep{de2021editing,charakorn2026doctolora}. Compared with existing work centered on task-level results~\citep{charakorn2026doctolora,back2026loramemory}, FSMem characterizes this influence through two complementary behaviors: \textbf{identifying valid actions} and \textbf{predicting execution outcomes}, as illustrated in Figure~\ref{fig:framework}(c). These behaviors reflect the two aspects of task rules encoded by the transition function: when an action can be executed and how execution changes the state. They therefore connect the experience represented in training data to the subsequent behavior that rule internalization should support.

We instantiate FSMem in synthetic worlds, classical planning, and code execution (Section~\ref{sec:setup}), and study how training-data content and organization shape parametric memory. Our study of \textbf{action complexity} shows that, under matched training conditions, task rules can become harder to internalize when learned alongside more action types (Section~\ref{sec:action}). Expanding \textbf{trajectory breadth} generally supports more effective rule reuse without increasing the number of training steps (Section~\ref{sec:breadth}). Our study of \textbf{depth} reveals depth-specialized behavior under single-depth training: models tend to perform better at execution depths represented in training. Based on this observation, training on a cumulative corpus of short and long trajectories can outperform long-only one with fewer tokens (Section~\ref{sec:depth}).

\section{Method}
\label{sec:formulation}

Our goal is to understand how training-data content and organization affect memory formation in LLMs. To this end, we introduce \textbf{FSMem}, a state-transition framework for constructing training data. We first formalize task rules and execution trajectories and define action complexity, trajectory breadth, and depth (Section~\ref{sec:world}). Building on this formulation, we describe how to construct supervised training examples from execution records (Section~\ref{sec:experience}).
\subsection{A Finite-State Machine Framework}
\label{sec:world}
To study how training data supports the learning of reusable execution rules, we need to specify both the rules underlying the data and the particular executions recorded. We use a finite-state machine to make this execution process explicit: its transition function specifies action validity and state changes, while its trajectories provide records for constructing training examples. This representation allows us to vary the action types selected for supervision or, with the rules fixed, change the state–action coverage and depth of execution records.
\paragraph{Finite-State Worlds}
To construct execution records under specified rules, we model each task instance as a deterministic finite-state machine, with configurations as states and concrete actions as inputs. The task world is specified by
\begin{equation}
\Omega=(\Ent,\Act,\St,F),
\end{equation}
where $\Ent$ is a finite set of typed entities, $\Act$ contains the action types under study, $\St$ is the finite state space, and $F$ is the transition function. 

A state \(s\) assigns values to state variables describing entity attributes and relations. Let \(\mathcal V\) denote a finite set of these variables, each with a finite value domain \(\mathcal D_v\). The state space consists of admissible configurations:
\begin{equation}
    \mathcal S\subseteq\prod_{v\in\mathcal V}\mathcal D_v,
    \qquad s=(s_v)_{v\in\mathcal V}.    
\end{equation}

\paragraph{Actions and Transitions}
The action set \(\mathcal A\) specifies which operation types are represented in the training data. Instantiating these types with entity arguments yields concrete actions, whose validity and execution outcomes are determined by \(F\). This separation of action validity and state changes follows classical planning representations~\citep{fikes1971strips,ghallab1998pddl}.

\begin{definition}[Action Complexity]
    \label{action-complexity}
    Let \(\mathcal A=\{a_1,a_2,\ldots,a_A\}\) be the set of distinct action identifiers represented in the corpus. The action complexity is $A=|\mathcal A|$.
\end{definition}
In our framework, each action type provides a template for concrete actions that serve as inputs to the finite-state machine. Instantiating the same type with different entity arguments does not increase action complexity.
\begin{definition}[Concrete Action]
    \label{def:concrete-action}
    A concrete action \(a_i(\mathbf{x})\) is obtained by instantiating an action \(a_i\) with a type-compatible tuple of entities \(\mathbf{x}\) from \(\mathcal E\). Here \(a_i\in\mathcal A,\ 1\le i\le A\).
\end{definition}
Reusing task rules requires knowing when a concrete action can trigger a valid state transition. We express the requirements for such transitions as execution conditions.
\begin{definition}[Execution Condition]
    \label{def:execution-cond}
    An execution condition \(p(\mathbf x)\) is a Boolean expression involving state variables and entity arguments \(\mathbf x\). Its value in state \(s=(s_v)_{v\in\mathcal V}\) is
    \begin{equation}
        p(s,\mathbf x)=\left.p(\mathbf x)\right|_{v=s_v,\;v\in\mathcal V}\in\{0,1\}.
    \end{equation}
\end{definition}

\begin{definition}[Valid Action]
    \label{def:valid-action}
    Let \(\mathcal P_i\) be the execution conditions for action type \(a_i\), and define
    \begin{equation}
        g_i(s,\mathbf x)=\bigwedge_{p\in\mathcal P_i}p(s,\mathbf x).
    \end{equation}

    A concrete action \(a_i(\mathbf x)\) is valid in state \(s\) if and only if \(g_i(s,\mathbf x)=1\).
\end{definition}
Identifying valid actions means determining whether a transition is permitted. Reusing task rules also requires predicting the state reached after executing a valid action.
\begin{definition}[Execution Outcomes]
    \label{def:execution}
    The execution outcome of a valid action \(a_i(\mathbf x)\) in state \(s\) is the resulting state \(s'\) after executing that action.
\end{definition}
Consistent with our view of parametric memory, past experience should induce lasting changes in model parameters that influence subsequent behavior. Here, the rules to be internalized govern action validity and state changes, jointly represented by the transition function \(F\).
\begin{definition}[Transition Function]
    \label{def:transition-function}
    The transition function $F$ encodes task rules by mapping each concrete action and current state to execution outcomes: 
    \begin{equation}
    F(s,a_i(\mathbf x)) = \begin{cases} s', & g_i(s,\mathbf x)=1,\\ \bot, & g_i(s,\mathbf x)=0, \end{cases}
    \end{equation}
Here, \(s'\in\mathcal S\) is the execution outcome and \(\bot\)  denotes an invalid transition.
\end{definition}

\paragraph{Execution Trajectories} 
In a finite-state machine, a sequence of valid actions produces a trajectory of states. We record how these actions change entity attributes and relations, and use the resulting execution trajectories to construct training data.
\begin{definition}[Execution Trajectory]
    \label{def:traj}
    An execution trajectory of length $T_\xi$ is a sequence $\xi=(s_0,u_1,s_1,\ldots,u_{T_\xi},s_{T_\xi})$,
    where $s_0\in\mathcal S$ and, for $1\le t\le T_\xi$, $u_t=a_{k_t}(\mathbf x_t)$ with $a_{k_t}\in\mathcal A$, and $s_t=F(s_{t-1},u_t)\in\mathcal S$.
\end{definition}
Training trajectories can follow the same task rules while differing in state–action coverage (\textbf{breadth}) and execution length (\textbf{depth}). These dimensions let us study how the organization of execution records shapes memory formation.
\begin{definition}[Trajectory Breadth]
    \label{def:breadth}
    Let \(\mathcal T\) be a collection of training trajectories. With the action set \(\mathcal A\), transition function \(F\), and trajectory depth fixed, breadth is the range of state-action pairs represented by these trajectories. For \(t\geq1\), their coverage at execution step \(t\) is
    \begin{equation}
        b_t(\mathcal T)=\left|\left\{\bigl(s_{t-1}^{(\xi)},u_t^{(\xi)}\bigr)\;\middle|\;\xi\in\mathcal T,\ T_\xi\geq t\right\}\right|.
    \end{equation}

    We expand breadth by collecting trajectories from additional distinct source examples.
\end{definition}

\begin{definition}[Trajectory Depth] 
    \label{def:depth}
    The depth of an execution trajectory \(\xi\) is the number of actions it contains: \(d(\xi)=T_\xi\).
\end{definition}

\Needspace{20\baselineskip}
\subsection{Constructing Training Data from Execution Records}
\label{sec:experience}

\begin{wrapfloat}{algorithm}{r}{0.50\textwidth}
\vspace{-\intextsep}
\begin{minipage}{\linewidth}
\small
\setlength{\parskip}{0pt}
\algrenewcommand{\algorithmicindent}{0.8em}
\begin{algorithm}[H]
\caption{Training Corpus Construction}
\label{alg:training-corpus}
\begin{algorithmic}[1]
\Require Action set $\mathcal A$, transition function $F$, source examples $\{z_i\}_{i=1}^{N}$, trajectory depths $\mathcal L$.
\Ensure Training corpus $\mathcal D=\{(q_j,y_j)\}_{j=1}^{M}$.
\State $\mathcal D\gets[\,]$; $M\gets0$ \label{algline:initialize}
\For{$i=1,\ldots,N$} \label{algline:source}
    \For{each $T\in\mathcal L$} \label{algline:depth}
        \For{each $\xi\in\Call{Collect}{z_i;\mathcal A,F,T}$} \label{algline:collect}
            \For{each $q\in\Call{Queries}{z_i,\xi}$} \label{algline:queries}
                \State $M\gets M+1$; $q_M\gets q$ \label{algline:index}
                \State $y_M\gets\Call{Label}{q_M,\xi;F}$ \label{algline:label}
                \State $\Call{Append}{\mathcal D,(q_M,y_M)}$ \label{algline:append}
            \EndFor
        \EndFor
    \EndFor
\EndFor
\State \Return $\mathcal D$ \label{algline:return}
\end{algorithmic}
\end{algorithm}
\end{minipage}
\vspace{-\intextsep}
\end{wrapfloat}

Algorithm~\ref{alg:training-corpus} constructs supervised pairs $(q_j,y_j)$
from source examples $z_i$ and selected trajectory depths $\mathcal L$.
After initializing the corpus (Line~\ref{algline:initialize}), it iterates over
each source and depth (Lines~\ref{algline:source}--\ref{algline:depth}).
\textsc{Collect} (Line~\ref{algline:collect}) generates or selects $T$-step
trajectories using action types from $\mathcal A$, with transitions satisfying
$s_t=F(s_{t-1},u_t)\neq\bot$.

For each trajectory, \textsc{Queries} (Line~\ref{algline:queries}) constructs
prediction inputs from the source example and observed execution information.
Each query is indexed (Line~\ref{algline:index}) and paired with a target
computed by \textsc{Label} from the execution record and $F$
(Line~\ref{algline:label}). The pair is appended to the corpus
(Line~\ref{algline:append}), which is returned in Line~\ref{algline:return}.

\par
\ifnum\value{WF@wrappedlines}>1
  \vspace{\dimexpr\value{WF@wrappedlines}\baselineskip-\baselineskip\relax}
\fi
\WFclear

\section{Experimental Setup}
\label{sec:setup}
\label{sec:queries}

This section describes the task instantiations, training and evaluation procedures, and experimental design used to study memory formation with FSMem.

\paragraph{\textbf{Task instantiations.}}
We instantiate Algorithm~\ref{alg:training-corpus} in three task families,
constructing execution records and deriving training targets from their task rules.
\textbf{(1) Synthetic worlds.} We execute valid actions from sampled entity
configurations; execution conditions and resulting states provide targets
for valid-action identification and state prediction.
\textbf{(2) Classical planning.} We construct records from problem instances
in 13 ACPBench domains~\citep{kokel2025acpbench}. Domain rules determine
valid-action targets for query states, specified directly or through an
initial state and an action sequence.
\textbf{(3) Code execution.} Running CodeI/O programs~\citep{li2025codeio}
on concrete inputs yields operation sequences and outputs. Recorded
operations supervise next-operation prediction; outputs supervise output prediction from the program and input (Native), optionally
augmented with execution order without intermediate values (Explicit trace).
Further details are provided in Appendix~\ref{app:data-construction}.

\paragraph{\textbf{Training and evaluation.}}
We fine-tune Qwen2.5-0.5B-Instruct~\citep{qwen2025qwen25technicalreport} with LoRA~\citep{hu2021lora}, keeping the procedure fixed within each comparison.
We assess memory formation through two complementary behaviors: identifying valid actions and predicting execution outcomes on held-out sources without access to training records.
In \textbf{synthetic worlds}, we evaluate valid-action selection by order-robust accuracy and state prediction by exact match.
In \textbf{classical planning}, we use candidate selection (order-robust accuracy) or complete valid-action-set prediction (exact-set accuracy and action-set F1).
In \textbf{code execution}, we evaluate program-output prediction by numeric-equivalence accuracy and, as a complementary test of rule reuse, next-operation selection by order-robust accuracy, using repository-disjoint training and test splits.
Order-robust accuracy requires correct answers under every tested candidate ordering.
Appendix~\ref{app:additional-models} reports additional valid-action identification experiments with Qwen2.5-3B-Instruct, Qwen2.5-7B-Instruct, and Qwen3-8B~\citep{yang2025qwen3} in synthetic worlds and classical planning.

\paragraph{Experimental Design} 
Our experiments aim to investigate three research questions about how training-data content and organization shape memory formation. First, \textbf{how does action complexity affect the effectiveness of task-rule internalization?} Controlled comparisons introduce additional action types while keeping existing actions' execution conditions and effects unchanged (Section~\ref{sec:action}). With these rules fixed, the remaining studies focus on trajectory organization. Second, \textbf{does broader state–action coverage support memory formation more effectively than repeatedly learning from fewer sources?} Breadth experiments vary the number of distinct source examples while holding trajectory depth and the number of training steps fixed (Section~\ref{sec:breadth}). Third, \textbf{how does training trajectory depth affect generalization across execution depths, and can short and long trajectories complement each other?} Depth experiments compare training at individual depths with training on cumulative corpora containing both shorter and longer trajectories (Section~\ref{sec:depth}).

\section{Action Complexity Affects Memory Formation Effectiveness}
\label{sec:action}

We examine how action complexity affects the 
effectiveness of memory formation. Increasing the number of action types represented in training data expands the range of task rules the model must learn. We assess how effectively models internalize and reuse these rules through two complementary behaviors: \textbf{identifying valid actions} in synthetic worlds and classical planning (Section~\ref{sec:action-boundary}), and \textbf{predicting execution outcomes} in synthetic worlds, with an additional test of next-operation prediction in code (Section~\ref{sec:action-consequence}).

\begin{figure}[!b]
\centering
\includegraphics[width=\textwidth]{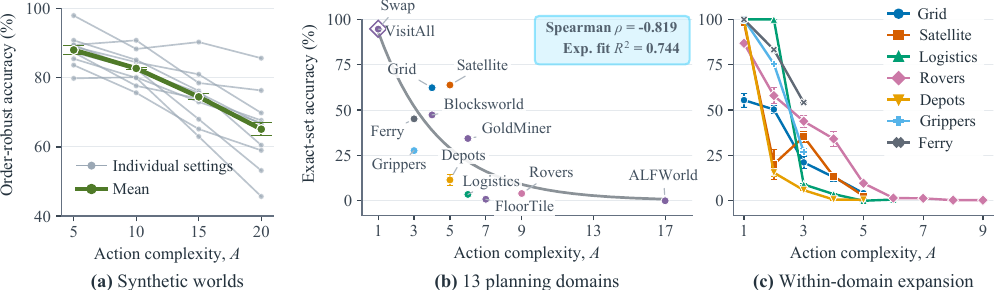}
\vspace{-6pt}
\caption{Valid-action identification as action complexity $A$ increases. \textbf{(a)} Synthetic order-robust accuracy: light curves show individual settings and the green curve their mean. \textbf{(b)} Exact-set accuracy in 13 planning domains; the gray curve is an exponential fit. \textbf{(c)} Exact-set accuracy as more action types are included in training within each of seven planning domains.}
\label{fig:action-boundary}
\end{figure}

\subsection{Identifying Valid Actions}
\label{sec:action-boundary}
We begin with synthetic worlds instantiated from our state-transition formulation, which allows us to vary action complexity in a controlled setting.
In synthetic worlds, we expand nested sets of action types from $A{=}5$ to $A{=}20$, holding the other world properties fixed within each comparison. Mean order-robust accuracy decreases from $88.1\%$ to $65.1\%$ (Figure~\ref{fig:action-boundary} a) with the same number of training steps and examples processed. These results indicate that increasing action complexity reduces the effectiveness of rule internalization under the same training conditions.
We next examine whether this finding could be extended to classical planning tasks.

In 13 planning domains, exact-set accuracy is negatively associated with the number of action types appearing in the ground-truth valid-action sets of the training data (Spearman $\rho=-0.819$; Figure~\ref{fig:action-boundary} b). We then progressively expand the set of action types within seven planning domains to examine whether the same decline appears (Figure~\ref{fig:action-boundary} c). Including more action types generally makes valid-action identification harder to learn within each domain.

\Needspace{18\baselineskip}
\subsection{Predicting Execution Outcomes}
\label{sec:action-consequence}

\begin{wrapfigure}{r}{0.34\textwidth}
\centering
\setlength{\abovecaptionskip}{2pt}
\setlength{\belowcaptionskip}{0pt}
\includegraphics[width=\linewidth]{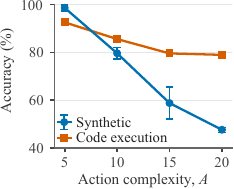}
\vspace{-14pt}
\caption{Next-state and next-operation prediction as action complexity increases.}
\label{fig:action-consequence}
\vspace{-6pt}
\end{wrapfigure}

Identifying valid actions tests whether a model has learned when an operation can be executed, but does not establish whether it can predict how execution proceeds. We examine next-state prediction in synthetic worlds and, as a complementary test of rule reuse, next-operation prediction in code.

In synthetic worlds, models predict complete next states on the same 200 state-action queries involving five shared action types. Exact accuracy decreases from \(98.7\%\) at \(A=5\) to \(47.7\%\) at \(A=20\) (Figure~\ref{fig:action-consequence}). A similar decline appears in code next-operation prediction. Given a program, its input, and an execution prefix without intermediate values, the model selects the next operation from two candidates. We expand the training action vocabulary from \(A=5\) to \(A=20\), keeping the same 800 training programs and evaluating the same 500 test events under both candidate orders. In this single-seed comparison, order-robust accuracy decreases from \(92.6\%\) to \(79.0\%\) (Figure~\ref{fig:action-consequence}). Even on identical questions about shared rules, models trained on a broader rule set make less reliable execution predictions.

\par
\makeatletter
\ifnum\c@WF@wrappedlines>1
  \vspace{\dimexpr\c@WF@wrappedlines\baselineskip-\baselineskip\relax}
\fi
\makeatother
\WFclear

\begin{takeaway}
Increasing action complexity impairs valid-action identification and execution-outcome prediction. The same task rules can become harder to internalize when learned alongside a wider range of actions.
\end{takeaway}

\section{Scaling Breadth to Improve Memory Formation}
\label{sec:breadth}
Learning task rules from execution trajectories requires generalizing beyond the particular entities and states recorded during training. To test whether broader state–action coverage supports this generalization, we vary breadth while holding action complexity and trajectory depth fixed (Section~\ref{sec:breadth-setting}), and assess valid-action identification (Section~\ref{sec:breadth-boundary}) and execution-outcome prediction (Section~\ref{sec:breadth-consequence}).

\subsection{Varying Breadth}
\label{sec:breadth-setting}

Following Algorithm~\ref{alg:training-corpus}, we collect trajectories \(\mathcal T\) from nested sets of \(N\) distinct source examples, with the action set \(\mathcal A\), transition function \(F\), and trajectory depths fixed. Increasing \(N\) adds trajectories in which the same action types are applied to different entities and states, expanding the state–action coverage \(b_t(\mathcal T)\) at each execution step \(t\) (synthetic counts in Table~\ref{tab:measured-breadth}). Within each task family, all conditions use the same number of training steps and effective batch size and share the same test set. \rev{At this fixed budget, increasing \(N\) spreads example presentations across more distinct sources, with fewer presentations per source. We test how this allocation affects rule reuse.}

\subsection{Trajectory Breadth Supports Valid-Action Identification}
\label{sec:breadth-boundary}
In synthetic worlds, increasing \(N\) from 50 to 1,600 raises order-robust accuracy from \(39.0\%\) to \(61.8\%\), with most of the gain reached by \(N=800\) (Figure~\ref{fig:breadth-applicability}, left): at fixed action complexity, broader state–action coverage helps the model learn the execution conditions of the same actions.

In classical planning, 12 of the 13 domains achieve higher exact-set accuracy at \(N=1{,}600\) than at \(N=50\), including gains of 20.1 percentage points on Grid and 18.0 points on Blocksworld (Figure~\ref{fig:breadth-applicability}, right); gains vary across domains, and ALFWorld remains near zero throughout the sweep. Together with Section~\ref{sec:action-boundary}, these results identify breadth as a way to address the difficulty of rule internalization without changing action complexity.

\begin{figurehere}
\begin{minipage}[t]{0.56\linewidth}
  \setlength{\abovecaptionskip}{0pt}
  \centering\vspace{0pt}
  \includegraphics[width=\linewidth]{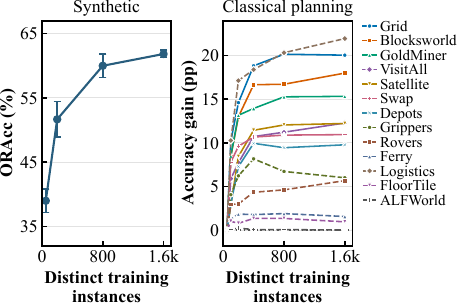}
  \vspace{-6pt}
  \begin{minipage}[t]{0.97\linewidth}
  \caption{\raggedright ORAcc: order-robust accuracy (mean $\pm$ SD). Planning: gains over $N{=}50$.}
  \label{fig:breadth-applicability}
  \end{minipage}
\end{minipage}\hfill
\begin{minipage}[t]{0.437\linewidth}
  \setlength{\abovecaptionskip}{0pt}
  \centering\vspace{0pt}
  \includegraphics[width=\linewidth]{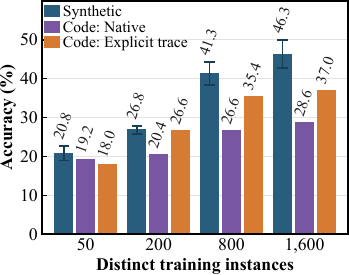}
  \vspace{-6pt}
  \begin{minipage}[t]{\linewidth}
    \caption{Next-state exact match and numeric-equiva\-lence output accuracy.}
  \label{fig:breadth-consequence}
  \end{minipage}
\end{minipage}
\end{figurehere}

\subsection{Breadth Scaling Enhances Execution-Outcome Prediction}
\label{sec:breadth-consequence}

For execution-outcome prediction, synthetic models predict complete next
states under the same $A{=}20$ rule set, and code models predict outputs on
the same 500 repository-disjoint test programs, trained on nested program
sets with a fixed mixture of short, medium, and long trajectories (S+M+L;
depth ranges defined in Section~\ref{sec:depth-consequence}).
Synthetic training uses 667,200 prompt tokens and 48,000 supervised tokens
per condition; code token counts are not exactly matched. Synthetic
next-state exact-match accuracy rises from $20.83\mathbin{\pm}1.89\%$ at
$N{=}50$ to $46.33\mathbin{\pm}3.55\%$ at $N{=}1{,}600$ (+25.5 points), and
program-output accuracy rises from $19.2\%$ to $28.6\%$ in the Native view
(+9.4 points) and from $18.0\%$ to $37.0\%$ in the Explicit-trace view
(+19.0 points) (Figure~\ref{fig:breadth-consequence}).

\par\WFclear
\begin{takeaway}
Expanding trajectory breadth generally improves memory formation, providing a way to mitigate the difficulty of rule internalization under higher action complexity. \end{takeaway}

\section{Benefits of Accumulating Trajectories across Depths}
\label{sec:depth}
Entity attributes and relations change along a trajectory, but the transition function governing an action does not change with depth, so a model that has learned the rules should reuse them at different trajectory \textbf{depths}. We first test this expectation and find that single-depth training produces depth-specialized behavior (Section~\ref{sec:depth-setting}). This motivates accumulating trajectories across depths while retaining shorter ones; we compare the resulting cumulative corpora with long-only training on valid-action identification (Section~\ref{sec:depth-boundary}) and execution-outcome prediction (Section~\ref{sec:depth-consequence}).

\vspace{-6pt}
\subsection{Single-Depth Training Produces Specialized Behavior}
\label{sec:depth-setting}

Motivated by known limits of length generalization~\citep{anil2022exploring},
we train separate models at individual depths and evaluate each across
depths in the synthetic, planning, and code task families.
Table~\ref{tab:depth-matching} reports, for selected settings, the matching
gain (the difference in mean accuracy between matched and mismatched
training--test depths) and $\rho$ (Spearman's correlation between the binary
depth-matching indicator and accuracy); matched-depth training achieves
higher accuracy on average in each. Figure~\ref{fig:depth-specialization}
shows the synthetic $A{=}5$ setting, where the matching gain is 12.2
percentage points and is most pronounced at shallow evaluation depths. Longer training executions thus do not automatically support stronger
performance at shorter depths.

\par\WFclear
\subsection{Valid-Action Identification across Depths}
\label{sec:depth-boundary}

\begin{figure*}[h!]
\vspace{-0.2cm}
    \begin{minipage}[t]{0.30\linewidth}
        \vspace{0pt}\centering
        \begingroup
  \makeatletter\def\@captype{table}\makeatother
  \centering
  \scriptsize
  \setlength{\tabcolsep}{2pt}
  \renewcommand{\arraystretch}{0.95}
  \caption{Depth matching in selected settings.}
  \label{tab:depth-matching}
  \resizebox{0.8\textwidth}{!}{\begin{tabular}{@{}llrr@{}}
    \toprule
    Task & Setting & Gain (pp) & $\rho$ \\
    \midrule
    Synthetic & $A=5$  & +12.2 & 0.751 \\
              & $A=20$ & +5.5 & 0.408 \\
    \midrule
    Planning & Ferry     & +20.1 & 0.836 \\
             & GoldMiner & +14.2 & 0.825 \\
             & Satellite & +38.2 & 0.825 \\
    \midrule
    Code & Native         & +14.8 & 0.504 \\
         & Explicit trace & +12.1 & 0.275 \\
    \bottomrule
  \end{tabular}}
\endgroup

    \end{minipage}\hfill
    \begin{minipage}[t]{0.28\linewidth}
        \vspace{0pt}\centering
        \includegraphics[width=0.9\linewidth]{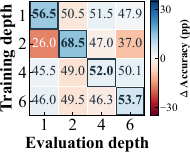}
        \caption{Synthetic depth specialization.}
        \label{fig:depth-specialization}
    \end{minipage}\hfill
    \begin{minipage}[t]{0.40\linewidth}
        \vspace{0pt}\centering
        \includegraphics[width=0.9\linewidth]{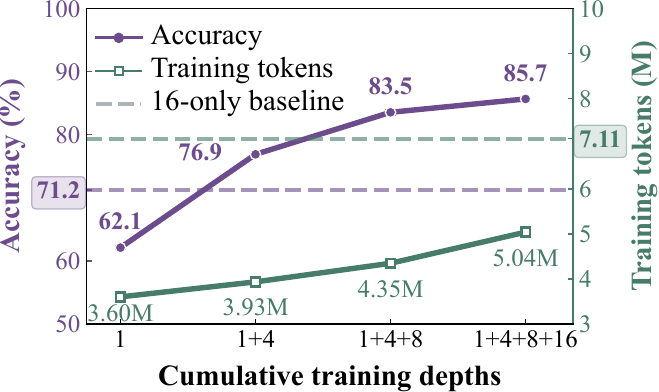}
        \caption{Valid-action identification.}
        \label{fig:depth-boundary-synthetic}
    \end{minipage}
\vspace{-0.3cm}
\end{figure*}

We test whether retaining shorter trajectories improves valid-action
identification as longer trajectories are added. The synthetic corpora cover
depths 1, $1{+}4$, $1{+}4{+}8$, or $1{+}4{+}8{+}16$, where ``$+$'' denotes
depths included together in one corpus, and 16-only training uses the same
maximum depth without shorter trajectories. The synthetic and code
conditions in Sections~\ref{sec:depth-boundary} and \ref{sec:depth-consequence}
each use 600 optimizer updates and 9,600 training examples.
Order-robust accuracy averaged over evaluation depths rises from $62.1\%$
to $85.7\%$ as depths are added, against $71.2\%$ for 16-only training
(Figure~\ref{fig:depth-boundary-synthetic}): a 14.5-point advantage with
29.1\% fewer non-padding tokens.

\begin{wraptable}{r}{0.62\textwidth}
  \vspace{-\intextsep}
  \centering
  \caption{\rev{Planning exact-set accuracy (\%) and token use.}}
  \label{tab:planning-depth-applicability}
  {\fontsize{7}{8.5}\selectfont
  \setlength{\tabcolsep}{1.1pt}
  \renewcommand{\arraystretch}{0.9}
  \begin{tabular}{@{}l*{9}{r}|r@{}}
    \toprule
    Depth & BW & Dep. & Ferry & Grip. & Sat. & Grid & GM & Swap & VA & Avg. \\
    \midrule
    $6$-only & 35.2 & 6.4 & 50.2 & 22.5 & 55.3 & 36.2 & 69.1 & 35.9 & 75.3 & \rev{42.90} \\
    $2$ & 31.5 & 6.7 & 33.7 & 9.8 & 32.7 & 31.1 & 37.6 & 35.4 & 46.8 & \rev{29.47} \\
    $2{+}4$ & 62.3 & 11.4 & 55.8 & 24.0 & 66.0 & 45.2 & 77.3 & 91.8 & 79.3 & \rev{57.01} \\
    $2{+}4{+}6$ & \textbf{77.0} & \textbf{11.8} & \textbf{67.8} & \textbf{31.1} & \textbf{95.9} & \textbf{47.1} & \textbf{99.4} & \textbf{99.7} & \textbf{80.0} & \rev{\textbf{67.76}} \\
    \midrule
    $\Delta$Tok & $-2.9$ & $-3.5$ & $-3.4$ & $-4.3$ & $-4.9$ & $-3.2$ & $-4.3$ & $-2.6$ & $-5.6$ & \rev{$-3.79$} \\
    \bottomrule
  \end{tabular}
  \par\smallskip
  {\raggedright\fontsize{6.5}{7.5}\selectfont
  BW:~Blocksworld; Dep.:~Depots; Grip.:~Grippers;
  Sat.:~Satellite; GM:~GoldMiner; VA:~VisitAll.
  \rev{$\Delta$Tok (\%) vs.\ 6-only.}\par}}
\end{wraptable}

\rev{At 300 updates and 4,800 presentations, cumulative training raises
mean exact-set accuracy from $42.90\%$ to $67.76\%$ across the nine domains
shown in Table~\ref{tab:planning-depth-applicability}, with $3.79\%$ fewer
non-padding training tokens in total. Accuracy improves in all nine domains.}
Full exact-set results for all 13 domains are in Appendix~\ref{app:planning-depth}.

\par\WFclear

\subsection{Execution-Outcome Prediction across Depths}
\label{sec:depth-consequence}

In synthetic worlds with varying action complexity $A$, models receive a
held-out initial state and an executed action prefix and predict the
complete terminal state. As the corpus
expands from depth 1 to $1{+}4{+}8{+}16$, terminal-state exact-match accuracy
averaged over evaluation depths rises from $32.3\%$ to $99.0\%$, with
corresponding non-padding training tokens reported in
Table~\ref{tab:synthetic-depth-consequence}.

To determine whether this gain comes only from the depth-16 trajectories,
we compare the full cumulative corpus with 16-only training at each
evaluation depth (Figure~\ref{fig:depth-consequence-synthetic}). The
cumulative corpus exceeds 16-only training throughout the tested synthetic
settings, $99.0\%$ against $64.4\%$ on average ($+34.6$ points) with 30.3\%
fewer non-padding tokens; the gap is concentrated at shorter depths (59.4,
62.2, and 16.5 points at depths 1, 4, and 8) and narrows to $+0.3$ points
at depth 16.

For program-output prediction, we construct corpora from the same 800
training programs using short (S: $T\leq6$), medium (M: $8\leq T\leq11$),
and long (L: $T\geq14$) trajectories. We evaluate cumulative corpora S,
S+M, and S+M+L on long executions against long-only (L-only) training,
shown as dashed baselines in Figure~\ref{fig:code-depth-consequence}.
The full cumulative corpus exceeds L-only training by 1.73 percentage
points in the Native view and 7.87 points in the Explicit-trace view, using
1.0\% and 15.1\% fewer non-padding tokens, respectively.
The benefit thus extends to
organizing execution records across depths within a fixed set of training
programs.

\par\medskip\noindent\begin{minipage}{\textwidth}
\begin{minipage}[t]{0.33\textwidth}
\vspace{0pt}\centering
\makeatletter\def\@captype{figure}\makeatother
\setlength{\abovecaptionskip}{0pt}
\includegraphics[width=\linewidth]{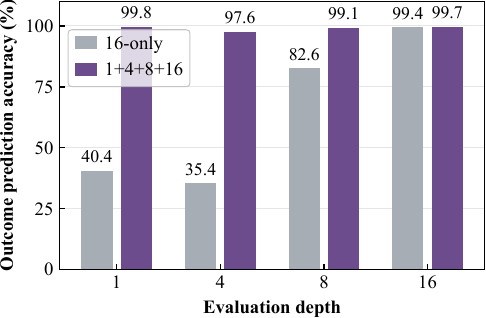}
\vspace{-8pt}
\caption{\raggedright Synthetic accuracy by evaluation depth.}
\label{fig:depth-consequence-synthetic}
\end{minipage}\hfill
\begin{minipage}[t]{0.28\textwidth}
\vspace{0pt}\centering
\begingroup
\makeatletter\def\@captype{table}\makeatother
\setlength{\abovecaptionskip}{0pt}
\setlength{\belowcaptionskip}{4pt}
\fontsize{8}{9.6}\selectfont
\setlength{\tabcolsep}{3pt}
\renewcommand{\arraystretch}{0.81}
\caption{\raggedright Synthetic Results.}
\label{tab:synthetic-depth-consequence}
\begin{tabular}{@{}lrr@{}}
\toprule
Training & Avg. acc. & Tokens \\
depths & (\%) & (M) \\
\midrule
16-only & 64.4 & 6.86 \\
1 & 32.3 & 3.34 \\
1+4 & 91.6 & 3.67 \\
1+4+8 & 98.3 & 4.09 \\
\textbf{1+4+8+16} & \textbf{99.0} & \textbf{4.78} \\
\bottomrule
\end{tabular}
\endgroup

\end{minipage}\hfill
\begin{minipage}[t]{0.36\textwidth}
\vspace{0pt}\centering
\makeatletter\def\@captype{figure}\makeatother
\setlength{\abovecaptionskip}{0pt}
\includegraphics[width=0.9\linewidth]{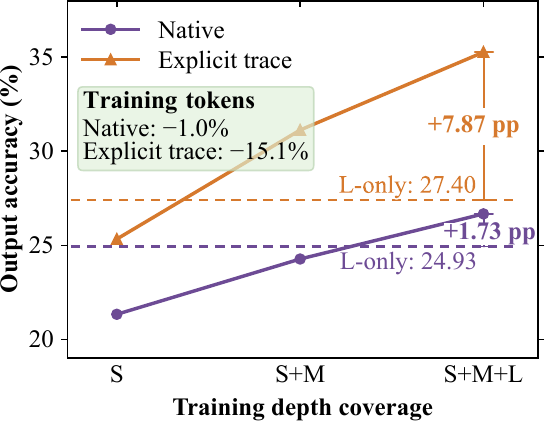}
\par
\caption{\raggedright Code Execution on long executions.}
\label{fig:code-depth-consequence}
\end{minipage}
\end{minipage}\par\medskip

\begin{takeaway}
Single-depth training produces depth-specialized behavior, while a cumulative corpus of short and long trajectories can outperform long-only training with fewer tokens.
\end{takeaway}

\section{Related Work}
\label{sec:related}

\paragraph{Parametric memory in LLMs.}
Analyses of pretrained Transformers characterize feed-forward layers as
key-value memories~\citep{geva2021transformer} and identify neurons associated
with particular facts~\citep{dai2022knowledge}.
Existing approaches update stored information through targeted parameter
changes~\citep{meng2022rome,fang2025alphaedit,wang2024wiserethinkingknowledgememory},
additional trainable parameters~\citep{dong2022calibrating,huang2023transformer}, or
dedicated memory modules~\citep{lample2019large,behrouz2025titans,wei2026mlpmemoryretrieverpretrainedmemory}.
Engram adds conditional memory through sparse lookup~\citep{cheng2026conditional},
while consolidation methods transfer short-term memories into long-term parameters
through distillation and replay~\citep{behrouz2026language}.
Recent LoRA-based studies examine capacity, retention, and reasoning over
internalized information~\citep{back2026loramemory,xu2026loraremembersparametricmemory}.

\paragraph{Training data and memory formation.}
Varied presentations of facts~\citep{allenzhu2024physics31}, supervision
formats~\citep{back2026loramemory}, and the number and frequency of
facts~\citep{ye2026cramless} influence acquisition and memorization.
For document-specific LoRA memory, training on question-answer pairs outperforms
next-token prediction on raw documents when answering questions about the
encoded content~\citep{liu2026shine}.
Data repetition also affects verbatim memorization~\citep{carlini2022quantifying,lee2022deduplicating}.
However, recalling stored information does not ensure its flexible use in
downstream reasoning~\citep{allen2025physics}.
Our study concerns execution records of rules shared across entities and
states. With fine-tuning fixed, we vary action complexity, trajectory breadth,
and depth to study how data construction supports rule internalization,
as reflected in identifying valid actions and predicting execution outcomes.

\paragraph{\rev{Generalization and transfer.}}
\rev{Compositional generalization tests whether models can recombine
familiar components in new ways~\citep{lake2018generalization,keysers2019measuring}.
Length generalization asks whether learned behavior extends to inputs longer
than those seen during training~\citep{anil2022exploring,zhou2024algorithms}.
Transfer learning studies how knowledge acquired in a source task or domain
helps in a target setting~\citep{pan2009survey,raffel2020exploring}.
FSMem studies how the content and organization of execution records shape
within-task generalization. We evaluate valid-action identification and
execution-outcome prediction on held-out instances governed by the same
task rules, with training records absent from the evaluation context.
Performance on these instances provides behavioral evidence of rule reuse.}

\section{Conclusion}
\label{sec:conclusion}

We introduced FSMem, a state-transition framework for studying how training-data content and organization shape parametric memory in LLMs. Our findings show that increasing action complexity can make even shared task rules harder to internalize, while broader state–action coverage supports more effective rule reuse. Single-depth training produces depth-specialized behavior, motivating cumulative corpora of short and long trajectories that can outperform long-only training with fewer tokens. Together, these findings show how training-data construction shapes the internalization of experience into reusable task rules, providing guidance for more effective memory formation under a fixed fine-tuning procedure.

\section*{AI Use Disclosure}
In this work, we used generative AI tools for generating synthetic
datasets through programmatic pipelines, helping develop the
conceptual framework, proposing or refining hypotheses, providing
feedback on research methodology and experiments, implementing
methods, assisting with translation, cleaning and reformatting
datasets, and interpreting results.
We have not used generative AI tools for formulating mathematical
claims, and providing critical ingredients for proving mathematical
claims, assisting in the writing of proofs, and supporting
qualitative and thematic data analysis are not applicable to this work.
Additionally, we used generative AI tools for creating or modifying
scientific figures, creating or editing software code, drafting
parts of the paper, summarizing and analyzing existing literature,
brainstorming, identifying relevant literature, improving manuscript
readability, formatting references, suggesting a paper structure,
and proposing titles and keywords.
We have reviewed all AI-assisted work. Specifically, we checked
cited sources, verified reported results against experimental
outputs, and reviewed AI-assisted code, text, and figures.
We take responsibility for the final content of this work, including
text, claims or artifacts produced with the aid of generative AI.

\bibliography{refs}
\bibliographystyle{fsmem_refs}

\clearpage
\appendix
\begingroup
\setlength{\parskip}{2pt}
\setlength{\intextsep}{6pt}
\section{Full Planning Depth Results}
\label{app:planning-depth}

Table~\ref{tab:planning-depth-fullset} reports exact-set accuracy for all
13 planning domains. Each score averages evaluation depths 2, 4, and 6
and then seeds 13, 37, and 73. Training on depths $2{+}4{+}6$ outperforms
6-only training in ten domains, ties in Logistics, and is lower in Rovers
and FloorTile, while using fewer non-padding training tokens in every domain.
Logistics, FloorTile, and ALFWorld remain near the exact-set floor.
\rev{Table~\ref{tab:planning-depth-applicability} uses the same exact-set
metric and the corresponding non-padding training-token counts for its ten
displayed domains. It averages accuracy across domains and computes the
aggregate token change from the summed counts.}

\begin{table}[ht]
  \centering
  \small
  \renewcommand{\arraystretch}{0.90}
  \setlength{\tabcolsep}{3pt}
  \caption{Full planning depth results. Exact-set accuracy (\%) averages
  evaluation depths 2, 4, and 6.
  $C$ denotes training on depths $2{+}4{+}6$; $L$ denotes 6-only training.
  $\Delta$Acc and $\Delta$Tok compare $C$ with $L$.
  Tokens are mean non-padding training tokens in millions.}
  \label{tab:planning-depth-fullset}
  \begin{tabular*}{\textwidth}{@{\extracolsep{\fill}}lrrrrrrrr@{}}
    \toprule
    & \multicolumn{5}{c}{Exact-set accuracy (\%)}
    & \multicolumn{3}{c}{Training tokens} \\
    \cmidrule(lr){2-6}\cmidrule(lr){7-9}
    Domain & $2$ & $2{+}4$ & $C$ & $L$ & $\Delta$Acc (pp)
      & $C$ (M) & $L$ (M) & $\Delta$Tok (\%) \\
    \midrule
    Blocksworld & 31.50 & 62.28 & 77.00 & 35.17 & +41.83 & 1.424 & 1.466 & $-2.9$ \\
    Depots & 6.72 & 11.44 & 11.83 & 6.39 & +5.44 & 2.972 & 3.080 & $-3.5$ \\
    Ferry & 33.72 & 55.78 & 67.78 & 50.17 & +17.61 & 2.342 & 2.424 & $-3.4$ \\
    FloorTile & 0.11 & 0.06 & 0.00 & 0.11 & $-0.11$ & 5.142 & 5.212 & $-1.3$ \\
    GoldMiner & 37.56 & 77.28 & 99.44 & 69.06 & +30.39 & 3.953 & 4.132 & $-4.3$ \\
    Grid & 31.06 & 45.22 & 47.11 & 36.22 & +10.89 & 3.911 & 4.039 & $-3.2$ \\
    Grippers & 9.78 & 24.00 & 31.11 & 22.50 & +8.61 & 2.239 & 2.341 & $-4.3$ \\
    Logistics & 1.39 & 0.56 & 0.67 & 0.67 & 0.00 & 4.250 & 4.412 & $-3.7$ \\
    Rovers & 5.06 & 5.61 & 6.28 & 6.61 & $-0.33$ & 4.626 & 4.765 & $-2.9$ \\
    Satellite & 32.67 & 66.00 & 95.89 & 55.33 & +40.56 & 2.527 & 2.656 & $-4.9$ \\
    Swap & 35.44 & 91.83 & 99.67 & 35.89 & +63.78 & 3.917 & 4.022 & $-2.6$ \\
    VisitAll & 46.78 & 79.28 & 80.00 & 75.33 & +4.67 & 1.988 & 2.106 & $-5.6$ \\
    ALFWorld & 0.22 & 0.17 & 0.06 & 0.00 & +0.06 & 28.429 & 28.772 & $-1.2$ \\
    \bottomrule
  \end{tabular*}
\end{table}

\section{Data Construction Details}
\label{app:data-construction}

\noindent\textbf{Synthetic worlds.}
Valid-action queries provide an initial state, an execution record, and
candidate actions; labels are obtained by checking execution conditions in
the resulting state. The action-complexity study of next-state prediction
instead supplies a complete current state and one valid concrete action.
Actions use fixed identifiers without descriptions of their effects, and
the target contains the resulting state of every entity, including entities
unchanged by the action. The depth study supplies an initial state and a
valid action sequence and uses the complete terminal state as the target.

\noindent\textbf{Classical planning.}
Problem instances specify objects and initial states. Domain rules generate
valid execution records and determine which concrete actions are valid at
each reached state. Current-state queries render the queried state directly;
trajectory queries provide the initial state and executed actions. Targets
either select a valid candidate or list the complete valid-action set.
Records derived from the same source problem remain in the same data split.
In the breadth experiments, Logistics and FloorTile use regenerated
controlled instances; the other planning domains use native-domain instance
construction.

\noindent\textbf{Code execution.}
Training and test programs are separated by repository. Next-operation
queries include the program description, input/output requirements, program,
concrete input, and an execution prefix with intermediate values removed.
The target operation is obtained from the recorded execution, and each
query is presented under both candidate orders. Program-output queries use
the same task information and concrete input; Explicit-trace inputs also
include executed line and statement order, without intermediate values or
the terminal output. Both Native and Explicit-trace inputs share the output
produced by execution as their target.

Distinct sources are sampled trajectories, planning problems, and programs
for synthetic worlds, classical planning, and code execution, respectively.
Breadth comparisons collect records from nested source pools with the action
types and trajectory depths fixed. Cumulative depth comparisons retain
records at shorter depths when records at longer depths are included.

\noindent\textbf{Result aggregation.}
Synthetic and planning breadth results average three training seeds; code
breadth results use seed 13. Synthetic cumulative-depth outcome results
also use seed 13, while the cumulative-depth code results average seeds
13, 37, and 73. Error bars report sample standard deviations where shown.

\subsection{Measured Trajectory Breadth}
\label{app:measured-breadth}
Table~\ref{tab:measured-breadth} reports $b_t(\mathcal T)$ from the frozen
synthetic training corpora, following Definition~\ref{def:breadth}.
Counts use symbolic states and executed concrete actions before surface
rendering; candidate-order variants and repeated training do not add pairs.
Increasing $N$ expands coverage at every step, although distinct trajectories
can share state--action pairs, so $b_t$ need not equal $N$.

\begin{table}[ht]
  \centering
  \small
  \setlength{\tabcolsep}{6pt}
  \begin{tabular*}{\linewidth}{@{\extracolsep{\fill}}lcrrrr@{}}
    \toprule
    Training corpus & Step $t$ & $N=50$ & $N=200$ & $N=800$ & $N=1600$ \\
    \midrule
    Valid-action identification (depth 4) & 1 & 50 & 193 & 683 & 1,191 \\
    & 2 & 49 & 180 & 602 & 982 \\
    & 3 & 47 & 149 & 399 & 604 \\
    & 4 & 35 & 97 & 192 & 291 \\
    \midrule
    Next-state prediction (depth 1) & 1 & 50 & 200 & 800 & 1,600 \\
    \bottomrule
  \end{tabular*}
  \caption{Measured breadth $b_t$ in synthetic training corpora ($A=20$); $N$ counts distinct sources.}
  \label{tab:measured-breadth}
\end{table}

\subsection{\rev{Synthetic Valid-Action Fine-Tuning}}
\label{app:synthetic-valid-action-finetuning}

\rev{Table~\ref{tab:synthetic-valid-action-finetuning} reports the LoRA
settings for Qwen2.5-0.5B-Instruct in the synthetic valid-action identification
experiment in Figure~\ref{fig:action-boundary}a.}

\begin{table}[htbp]
  \centering
  \small
  \caption{\rev{Fine-tuning settings for synthetic valid-action identification.}}
  \label{tab:synthetic-valid-action-finetuning}
  \begin{tabular}{@{}ll@{}}
    \toprule
    \rev{Setting} & \rev{Value} \\
    \midrule
    \rev{Optimizer} & \rev{AdamW} \\
    \rev{LoRA rank / alpha / dropout} & \rev{8 / 16 / 0} \\
    \rev{LoRA target modules} & \rev{\texttt{q\_proj}, \texttt{v\_proj}} \\
    \rev{Peak learning rate} & \rev{$2\times10^{-4}$} \\
    \rev{Example presentations per optimizer update} & \rev{16} \\
    \bottomrule
  \end{tabular}
\end{table}
\endgroup

\clearpage
\section{Results on Additional Models}
\label{app:additional-models}

We extend valid-action identification to Qwen2.5-3B-Instruct,
Qwen2.5-7B-Instruct, and Qwen3-8B. Candidate selection uses order-robust
accuracy, requiring correct answers under every tested ordering.
Completed comparisons retain all intermediate conditions.

\subsection{Action Complexity and Trajectory Breadth}

\begin{figure}[!htbp]
\centering
\includegraphics[width=0.96\linewidth]{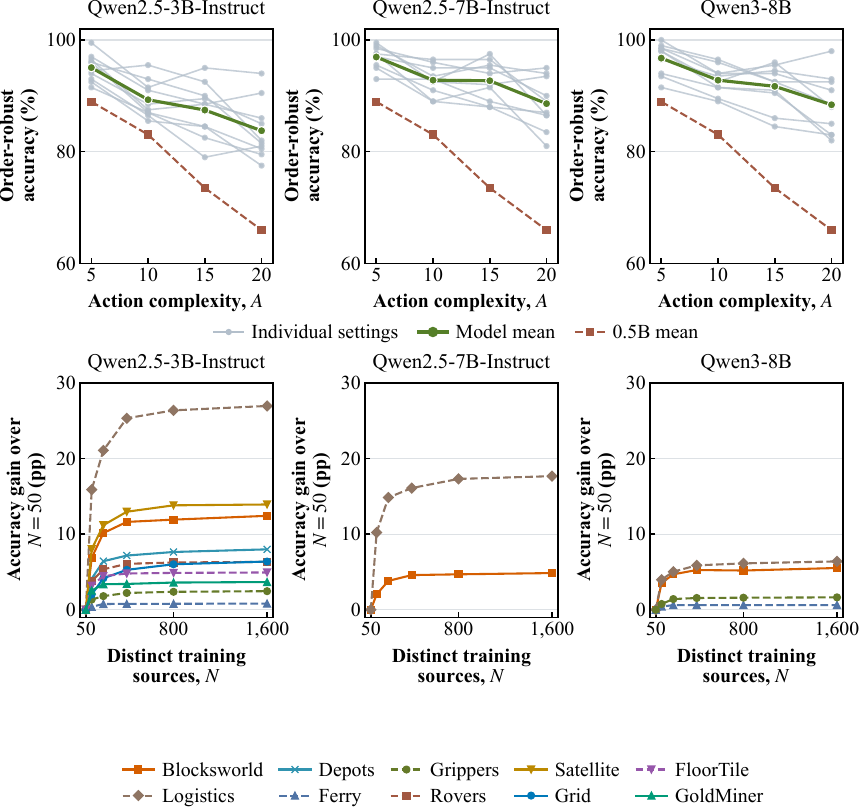}
\caption{Action complexity and breadth across models. Top: synthetic accuracy; gray lines show individual settings, green their mean, and dashed brown the 0.5B reference. Bottom: planning gains relative to $N=50$, where $N$ counts distinct training sources; colors identify domains.}
\label{fig:am-content-breadth}
\end{figure}

\paragraph{Action complexity.}
Across the same 40 synthetic configurations and 300 training steps
(seed 13), mean accuracy decreases from $A=5$ to $A=20$ in all three
models. The decline is smaller than for 0.5B: larger models mitigate,
but do not remove, the effect of action complexity in this comparison.

\paragraph{Trajectory breadth.}
Increasing distinct sources expands state--action coverage while action
complexity and trajectory depths remain fixed. Each condition uses 300
steps and 4,800 training examples. The $N=1600$ mean exceeds $N=50$ in
all 16 complete model--domain comparisons, despite intermediate plateaus
and decreases. Broader coverage remains beneficial across these models
without additional training steps.

\clearpage
\subsection{Depth Specialization and Cumulative Training}
\begin{figure}[!htbp]
\centering
\includegraphics[width=0.96\linewidth]{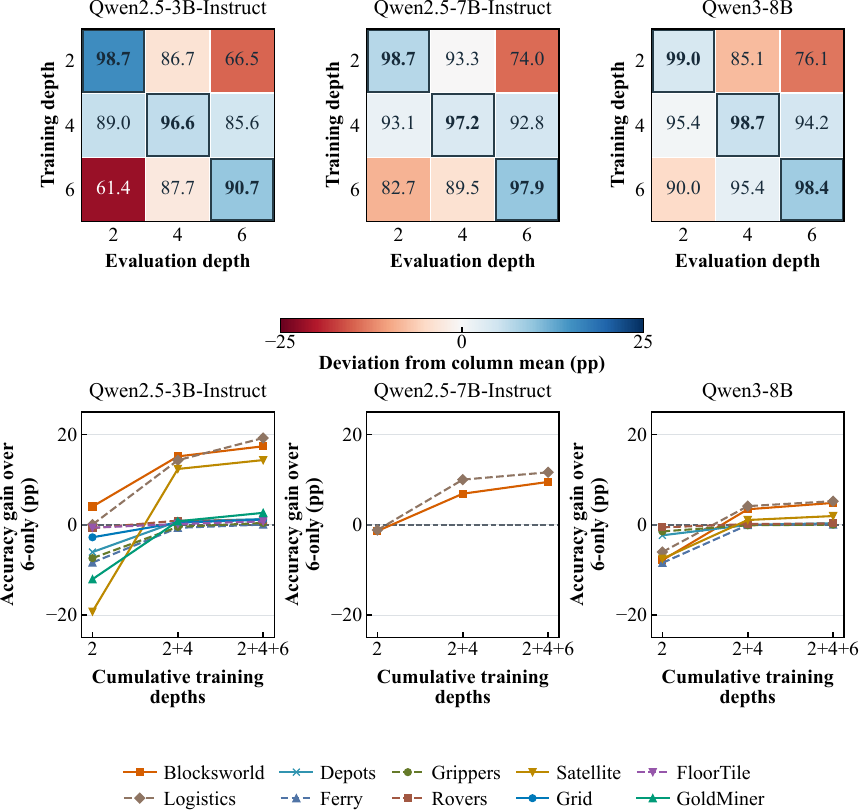}
\caption{Depth specialization and cumulative training. Top: Blocksworld training--test matrices; numbers show accuracy (\%), colors show deviations from column means, and outlines mark matched depths. Bottom: cumulative-training gains over 6-only training (dashed zero line). Each condition uses 300 steps; colors identify domains.}
\label{fig:am-depth}
\end{figure}

Figure~\ref{fig:am-depth} connects single-depth specialization with
cumulative training. Planning candidate-selection scores in this section average
four candidate counts ($2$, $3$, $4$, $8$), three training seeds (13, 37,
73), and three evaluation depths ($2$, $4$, $6$), except when these
factors are shown separately. Every required evaluation must be present.

\paragraph{Cumulative training.}
The $2{+}4{+}6$ corpus outperforms 6-only training in all 19 complete
model--domain comparisons. The plus sign denotes trajectory depths
included together in one corpus. Accuracy also increases from $2$ to
$2{+}4$ and then to $2{+}4{+}6$ in every comparison. Retaining shorter
trajectories therefore supports valid-action identification across
models without increasing training steps.

\clearpage
\paragraph{Training--test depth matching.}
Table~\ref{tab:am-depth-matrices} provides the other 22 complete matrices;
the three Blocksworld matrices appear in Figure~\ref{fig:am-depth}.
Each cell averages four candidate counts and three training seeds.
The matching gain, $\Delta$, is the mean of the three matched-depth cells
minus the mean of the six mismatched-depth cells. It is positive in all
25 matrices: ten for Qwen2.5-3B, eight for Qwen2.5-7B, and seven for
Qwen3-8B. This supports depth specialization as an average tendency;
the matched model need not be best at every test depth.

\begin{table}[!htbp]
\centering
\begingroup
\small
\setlength{\tabcolsep}{3pt}
\begin{tabular*}{\linewidth}{@{\extracolsep{\fill}}lrrrrrrrrrr@{}}
\toprule
& \multicolumn{3}{c}{Train depth 2} & \multicolumn{3}{c}{Train depth 4} & \multicolumn{3}{c}{Train depth 6} & \\
\cmidrule(lr){2-4}\cmidrule(lr){5-7}\cmidrule(lr){8-10}

Test depth & 2 & 4 & 6 & 2 & 4 & 6 & 2 & 4 & 6 & $\Delta$ \\

\midrule

\multicolumn{11}{l}{\textit{Qwen2.5-3B-Instruct}} \\

Logistics & \textbf{86.6} & 72.5 & 70.0 & 81.9 & \textbf{77.4} & 79.7 & 76.7 & 77.0 & \textbf{75.2} & +3.4 \\

Depots & \textbf{99.1} & 92.8 & 84.0 & 97.8 & \textbf{98.7} & 97.5 & 96.9 & 98.1 & \textbf{99.0} & +4.5 \\

Ferry & \textbf{100.0} & 95.0 & 79.6 & 99.9 & \textbf{99.8} & 97.9 & 99.8 & 100.0 & \textbf{100.0} & +4.6 \\

Grippers & \textbf{99.9} & 93.4 & 82.8 & 96.4 & \textbf{99.5} & 95.5 & 99.1 & 99.8 & \textbf{99.3} & +5.1 \\

Rovers & \textbf{99.9} & 99.2 & 94.6 & 98.1 & \textbf{99.0} & 97.9 & 98.1 & 98.8 & \textbf{98.9} & +1.5 \\

Satellite & \textbf{99.2} & 54.5 & 43.9 & 62.0 & \textbf{98.5} & 89.1 & 61.8 & 94.6 & \textbf{98.7} & +31.1 \\

Grid & \textbf{99.1} & 95.8 & 91.3 & 98.6 & \textbf{98.8} & 96.1 & 98.6 & 98.3 & \textbf{97.5} & +2.0 \\

FloorTile & \textbf{100.0} & 98.1 & 96.9 & 99.7 & \textbf{99.0} & 98.7 & 98.5 & 98.9 & \textbf{99.4} & +1.0 \\

GoldMiner & \textbf{100.0} & 91.0 & 64.7 & 99.6 & \textbf{99.8} & 93.9 & 94.4 & 97.8 & \textbf{99.8} & +9.6 \\

\addlinespace[3pt]

\multicolumn{11}{l}{\textit{Qwen2.5-7B-Instruct}} \\

Logistics & \textbf{92.5} & 80.7 & 77.7 & 90.9 & \textbf{89.0} & 87.4 & 83.6 & 84.6 & \textbf{86.0} & +5.0 \\

Depots & \textbf{99.9} & 98.2 & 96.6 & 99.7 & \textbf{99.5} & 99.3 & 99.2 & 98.9 & \textbf{99.5} & +1.0 \\

Ferry & \textbf{100.0} & 98.0 & 93.2 & 100.0 & \textbf{100.0} & 99.7 & 100.0 & 100.0 & \textbf{100.0} & +1.5 \\

Grippers & \textbf{100.0} & 98.0 & 94.7 & 98.3 & \textbf{99.6} & 95.5 & 100.0 & 100.0 & \textbf{99.7} & +2.0 \\

Satellite & \textbf{99.8} & 77.2 & 65.5 & 85.4 & \textbf{100.0} & 96.1 & 83.4 & 99.2 & \textbf{99.5} & +15.3 \\

Grid & \textbf{99.7} & 98.4 & 98.3 & 99.6 & \textbf{99.2} & 99.1 & 99.5 & 98.8 & \textbf{99.2} & +0.4 \\

FloorTile & \textbf{99.8} & 98.1 & 95.1 & 99.8 & \textbf{99.9} & 99.1 & 99.8 & 99.9 & \textbf{99.4} & +1.0 \\

\addlinespace[3pt]

\multicolumn{11}{l}{\textit{Qwen3-8B}} \\

Logistics & \textbf{98.0} & 83.3 & 77.8 & 96.7 & \textbf{91.8} & 89.0 & 92.5 & 93.2 & \textbf{91.5} & +5.0 \\

Depots & \textbf{99.5} & 97.0 & 94.9 & 99.5 & \textbf{99.6} & 99.1 & 99.5 & 99.5 & \textbf{99.5} & +1.3 \\

Ferry & \textbf{100.0} & 91.1 & 83.4 & 100.0 & \textbf{100.0} & 100.0 & 100.0 & 99.9 & \textbf{100.0} & +4.3 \\

Grippers & \textbf{99.9} & 99.0 & 96.4 & 99.7 & \textbf{99.9} & 99.3 & 99.9 & 100.0 & \textbf{99.8} & +0.8 \\

Rovers & \textbf{99.8} & 99.1 & 97.7 & 99.5 & \textbf{99.5} & 99.0 & 99.7 & 99.4 & \textbf{99.2} & +0.5 \\

Satellite & \textbf{99.8} & 85.8 & 85.6 & 90.4 & \textbf{99.6} & 95.7 & 94.5 & 99.4 & \textbf{99.8} & +7.9 \\

\bottomrule
\end{tabular*}
\endgroup
\caption{Training--test depth matrices for the remaining planning domains. Values are order-robust accuracy (\%); bold marks matched depths. $\Delta$ is the matching gain in percentage points. Blocksworld appears in Figure~\ref{fig:am-depth}.}
\label{tab:am-depth-matrices}
\end{table}

\paragraph{Scope.}
These experiments extend the valid-action findings to additional Qwen
models. They do not evaluate execution-outcome prediction or establish
token savings for these models. Breadth is varied through source-pool
size; $N$ is not a count of distinct state--action pairs. The comparisons
support the effects of data construction within each model; differences
between Qwen2.5 and Qwen3 also involve model generation, not only size.

\clearpage
\subsection{Results by Candidate Count}
\begin{figure}[!htbp]
\centering
\includegraphics[width=0.96\linewidth]{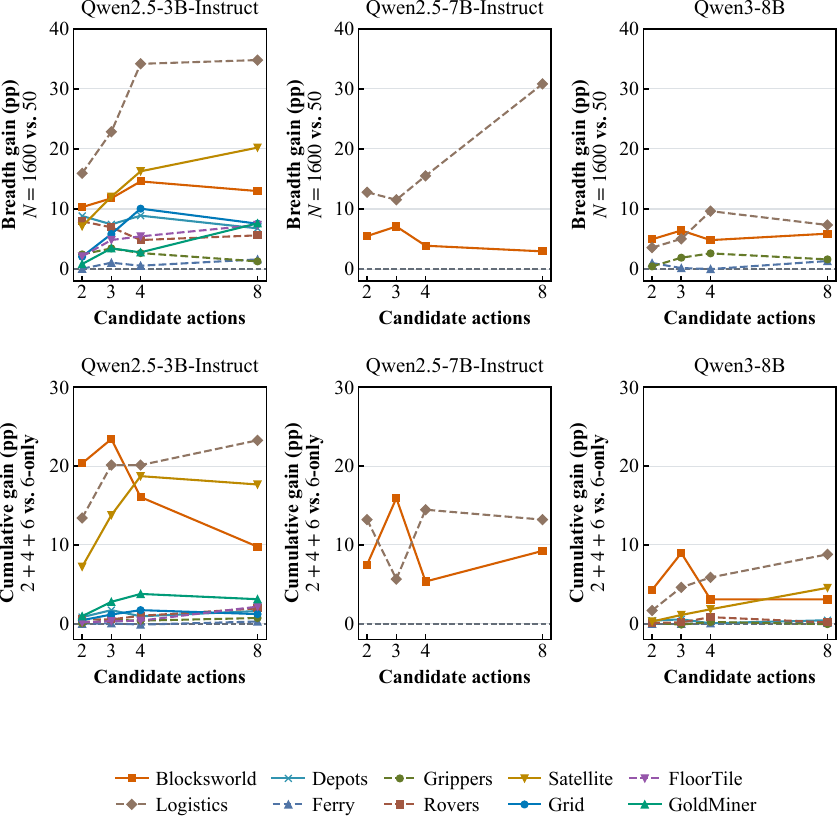}
\caption{Planning gains by candidate count. Top: $N=1600$ minus $N=50$. Bottom: cumulative $2{+}4{+}6$ minus 6-only training. All comparisons use 300 training steps. The dashed line marks zero gain; colors identify domains.}
\label{fig:am-candidates}
\end{figure}

Figure~\ref{fig:am-candidates} separates the planning results by candidate
count to check whether the benefits depend on tasks with few candidates.
All 16 complete breadth groups and all 19 complete cumulative-depth
groups are retained. Each point averages three training seeds and three
evaluation depths. These are disaggregated results from the same
experiments, rather than additional training runs.

Breadth yields positive gains in 63 of 64 comparisons and ties in one.
Cumulative training improves 70 of 76 comparisons, ties in three, and
decreases slightly in three (at most 0.12 percentage points).
Both gains remain positive at every candidate count in Blocksworld and
Logistics for all three models, including the eight-candidate setting.
The average benefits therefore do not arise only from tasks with few
candidates.

\clearpage
\subsection{Cumulative Training at Different Evaluation Depths}
\begin{figure}[!htbp]
\centering
\includegraphics[width=0.96\linewidth]{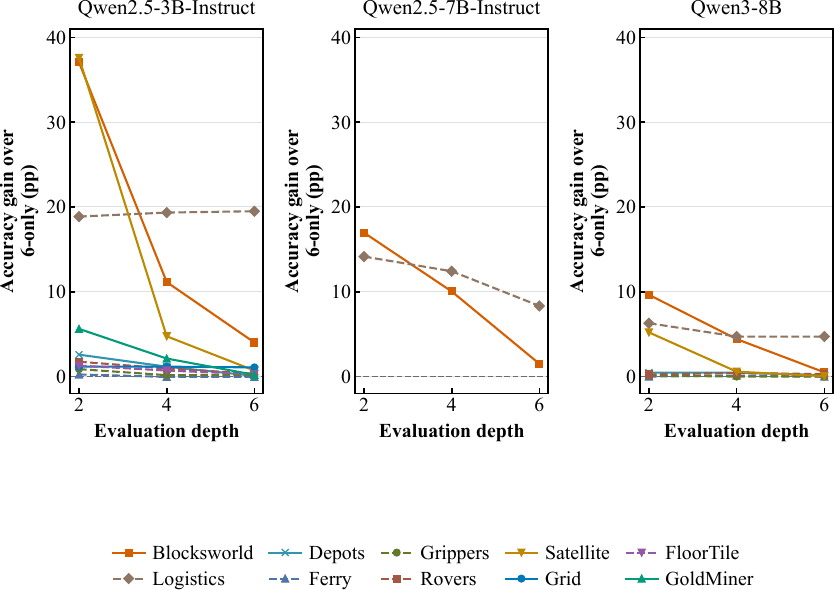}
\caption{Cumulative-training gains at each evaluation depth. Curves show $2{+}4{+}6$ minus 6-only order-robust accuracy in all complete planning comparisons. The dashed line marks zero gain; colors identify domains.}
\label{fig:am-test-depth}
\end{figure}

The preceding cumulative-training curves average over test depths.
Figure~\ref{fig:am-test-depth} separates these depths to locate the gains
from retaining shorter trajectories. It compares $2{+}4{+}6$ with 6-only
training in all 19 complete model--domain groups, using 300 training steps
per condition. Each point averages four candidate counts and three
training seeds.

Cumulative training improves 18 of 19 comparisons at depth 2, 17 at depth
4, and 16 at depth 6. The remaining comparisons comprise four ties and
two decreases smaller than 0.05 percentage points. Within each model's
completed task set, the mean gain is greatest at depth 2 and remains
positive at depth 6. Retaining shorter trajectories therefore helps
most at shallower depths on average, while often also benefiting the
deepest tested executions.

These are the same trained models and evaluation records underlying
Figure~\ref{fig:am-depth}, separated by test depth. Model columns contain
different sets of completed domains and should not be compared as a
controlled model-size sweep.

\clearpage
\subsection{Additional Planning Comparisons}
\paragraph{Complete valid-action sets.}
Table~\ref{tab:am-planning-exactset} complements candidate selection by
requiring the entire valid-action set to be correct. All three models
have results for 11 domains. Training uses 800 source instances for three
epochs; scores average seeds 1, 2, and 3 on the original combined test set.

\begin{table}[!htbp]
\centering
\begingroup\small
\renewcommand{\arraystretch}{0.90}
\setlength{\tabcolsep}{3pt}
\begin{tabular*}{\linewidth}{@{\extracolsep{\fill}}lrrr@{}}
\toprule

Domain & Qwen2.5-3B & Qwen2.5-7B & Qwen3-8B \\

\midrule

Blocksworld & 68.00 & 88.17 & 86.00 \\

Depots & 56.00 & 65.00 & 68.33 \\

Ferry & 67.83 & 92.00 & 92.33 \\

FloorTile & 0.50 & 0.00 & 4.17 \\

GoldMiner & 47.83 & 71.83 & 64.33 \\

Grid & 67.00 & 71.17 & 69.83 \\

VisitAll & 93.50 & 97.17 & 97.17 \\

Grippers & 51.50 & 85.67 & 91.00 \\

Logistics & 5.83 & 22.67 & 18.00 \\

Rovers & 7.83 & 12.33 & 18.00 \\

Satellite & 80.00 & 88.67 & 89.67 \\

\bottomrule
\end{tabular*}
\endgroup
\caption{Complete valid-action-set accuracy (\%) in 11 planning domains.}
\label{tab:am-planning-exactset}
\end{table}

\paragraph{Training-step sensitivity.}
Table~\ref{tab:am-additional-steps} compares 300 total steps with 300
steps per included depth: 600 for $2{+}4$ and 900 for $2{+}4{+}6$.
Additional steps improve 32 of 38 comparisons, tie in three, and reduce
accuracy in three. Cumulative training can therefore retain room for
further optimization. Its advantage over long-only training in
Figure~\ref{fig:am-depth} is established separately at matched step counts.

\begin{table}[!htbp]
\centering
\begingroup\small
\renewcommand{\arraystretch}{0.90}
\setlength{\tabcolsep}{3pt}
\begin{tabular*}{\linewidth}{@{\extracolsep{\fill}}lrrrr@{}}
\toprule

& \multicolumn{2}{c}{$2{+}4$} & \multicolumn{2}{c}{$2{+}4{+}6$} \\
\cmidrule(lr){2-3}\cmidrule(l){4-5}

Domain & 300 steps & 600 steps & 300 steps & 900 steps \\

\midrule

\multicolumn{5}{l}{\textit{Qwen2.5-3B-Instruct}} \\

Blocksworld & 95.08 & 96.15 & 97.32 & 98.03 \\

Logistics & 90.62 & 92.03 & 95.55 & 95.96 \\

Depots & 98.76 & 99.33 & 99.28 & 99.64 \\

Ferry & 99.24 & 99.39 & 99.96 & 99.96 \\

Grippers & 98.94 & 99.19 & 99.82 & 99.82 \\

Rovers & 99.56 & 99.69 & 99.64 & 99.82 \\

Satellite & 97.43 & 97.46 & 99.40 & 99.71 \\

Grid & 98.62 & 98.97 & 99.28 & 99.60 \\

FloorTile & 99.14 & 99.42 & 99.72 & 99.76 \\

GoldMiner & 98.08 & 98.39 & 99.94 & 99.97 \\

\addlinespace[3pt]

\multicolumn{5}{l}{\textit{Qwen2.5-7B-Instruct}} \\

Blocksworld & 96.92 & 97.56 & 99.53 & 99.33 \\

Logistics & 94.76 & 94.97 & 96.38 & 96.70 \\

\addlinespace[3pt]

\multicolumn{5}{l}{\textit{Qwen3-8B}} \\

Blocksworld & 98.03 & 98.29 & 99.43 & 99.63 \\

Logistics & 96.54 & 96.65 & 97.64 & 98.01 \\

Depots & 99.46 & 99.67 & 99.85 & 99.90 \\

Ferry & 99.92 & 99.92 & 99.99 & 100.00 \\

Grippers & 99.79 & 99.75 & 99.93 & 100.00 \\

Rovers & 99.60 & 99.51 & 99.75 & 99.79 \\

Satellite & 98.99 & 99.15 & 99.86 & 99.93 \\

\bottomrule
\end{tabular*}
\endgroup
\caption{Order-robust accuracy (\%) with 300 total steps or 300 steps per included depth.}
\label{tab:am-additional-steps}
\end{table}

\end{document}